\documentclass[sigconf]{acmart}

\usepackage{makecell}
\usepackage{wrapfig}
\usepackage{enumitem}  %
\usepackage{amsmath}   %
\usepackage{longtable}
\usepackage{booktabs}
\usepackage{multirow}
\usepackage{graphicx}
\usepackage{subcaption}
\usepackage{array}
\usepackage{tabularx}

\AtBeginDocument{%
  }

\setcopyright{acmlicensed}
\copyrightyear{2018}
\acmYear{2018}
\acmDOI{XXXXXXX.XXXXXXX}
\acmConference[Conference acronym 'XX]{Make sure to enter the correct
  conference title from your rights confirmation emai}{June 03--05,
  2018}{Woodstock, NY}
\acmISBN{978-1-4503-XXXX-X/18/06}

\newcommand{\systemname}{TrafficSimAgent}
\begin{document}

\title{\systemname: A Hierarchical Agent Framework for Autonomous Traffic Simulation with MCP Control}
\author{%
{ \bf Yuwei Du$^1$, Jun Zhang$^1$, Jie Feng$^1$, Zhicheng Liu$^2$, Jian Yuan$^1$, Yong Li$^1$ } \\
$^1$Department of Electronic Engineering, BNRist, Tsinghua University, Beijing, China\\
$^2$AMAP, Alibaba Group, Beijing, China \\
\texttt{fengj12ee@hotmail.com liyong07@tsinghua.edu.cn}
}

\renewcommand{\shortauthors}{Yuwei Du et al.}

\begin{abstract}
Traffic simulation is important for transportation optimization and policy making. While existing simulators such as SUMO and MATSim offer fully-featured platforms and utilities, users without too much knowledge about these platforms often face significant challenges when conducting experiments from scratch and applying them to their daily work. 
To solve this challenge, we propose \systemname, an LLM-based agent framework that serves as an expert in experiment design and decision optimization for general-purpose traffic simulation tasks. 
The framework facilitates execution through cross-level collaboration among expert agents: high-level expert agents comprehend natural language instructions with high flexibility, plan the overall experiment workflow, and invoke corresponding MCP-compatible tools on demand; meanwhile, low-level expert agents select optimal action plans for fundamental elements based on real-time traffic conditions. Extensive experiments across multiple scenarios show that \systemname~effectively executes simulations under various conditions and consistently produces reasonable outcomes even when user instructions are ambiguous. Besides, the carefully designed expert-level autonomous decision-driven optimization in \systemname~yields superior performance when compared with other systems and SOTA LLM based methods. 
\end{abstract}

\maketitle

\section{Introduction} \label{sec:introduction}

Traffic simulation is of paramount importance as a critical platform for optimizing transportation infrastructure development, refining traffic management policies, and enhancing overall travel efficiency~\cite{behrisch2011sumo,w2016multi}. Furthermore, traffic simulation plays a vital role across various interdisciplinary fields~\cite{zhang2024moss}, such as modeling dynamic human flow in urban planning, or simulating vehicle movement patterns for environmental monitoring and management. Given this significant value, numerous traffic simulation platforms have emerged, each designed for specific formatting and modeling needs, including tools like SUMO~\cite{behrisch2011sumo}, MATSim~\cite{w2016multi}, CityFlow~\cite{zhang2019cityflow}, and MOSS~\cite{zhang2024moss}. These platforms offer diverse tools to address and optimize various traffic-related challenges.

However, the effective use of these existing platforms relies heavily on expert knowledge~\cite{li2024chatsumo,ye2025sumo,w2016multi}. This dependence creates significant barriers for many users, particularly those from interdisciplinary backgrounds, resulting in several key difficulties: preparing simulation data is challenging, intervening in the simulation process is complex, and interpreting and optimizing the simulation results is often difficult. These issues severely limit the practical application of these platforms and hinder the effective optimization of real-world problems.

In recent years, the emergence of Large Language Models, endowed with vast common sense and powerful reasoning and planning capabilities, has made it possible to build automated workflows for complex real-world tasks. This approach has seen rapid development in fields such as WebAgent~\cite{zhou2023webarena,lai2024autowebglm} and CodingAgent~\cite{qian2023chatdev,yang2024swe}, with notable examples like WebArena~\cite{zhou2023webarena} and SWE-Agent~\cite{yang2024swe}. This success has inspired researchers in the transportation domain, leading to several new works, including TrajAgent~\cite{du2024trajagent}, ChatSUMO~\cite{li2024chatsumo}, and SUMO-MCP~\cite{ye2025sumo}. However, TrajAgent~\cite{du2024trajagent} focuses on modeling movement trajectories, such as mobility data generation, and does not directly support traffic simulation. While both ChatSUMO~\cite{li2024chatsumo} and SUMO-MCP~\cite{ye2025sumo} support SUMO-based traffic simulation, they are limited to the automated execution of a fixed set of scenarios and parameters. Crucially, they all lack systematic automatic optimization and feedback capabilities, preventing effective iterative improvement and optimization.

To address the aforementioned challenges, we propose \systemname, a novel LLM based multi-agent framework for autonomous traffic simulation. Specifically, \systemname~introduces three core components: first, it defines and abstracts API functions, decoupling them from the underlying traffic simulation platform and packaging them as MCP Functions to support automatic calling by the LLM Agents; second, it features a Task Understanding and Autonomous Planning module that automatically interprets vague natural language instructions, enabling the agent to freely combine underlying functional modules and break away from the fixed workflows, which significantly enhances the generality and adaptability of the automated simulation system to different scenarios; and finally, integrated with the simulation process, it includes a built-in system optimization module that supports both a full-stack automatic optimization driven entirely by LLM and a two-layer optimization logic combining the LLM's strengths with classical low-level optimization algorithms.

\begin{figure*}
    \centering
    \includegraphics[width=1\textwidth]{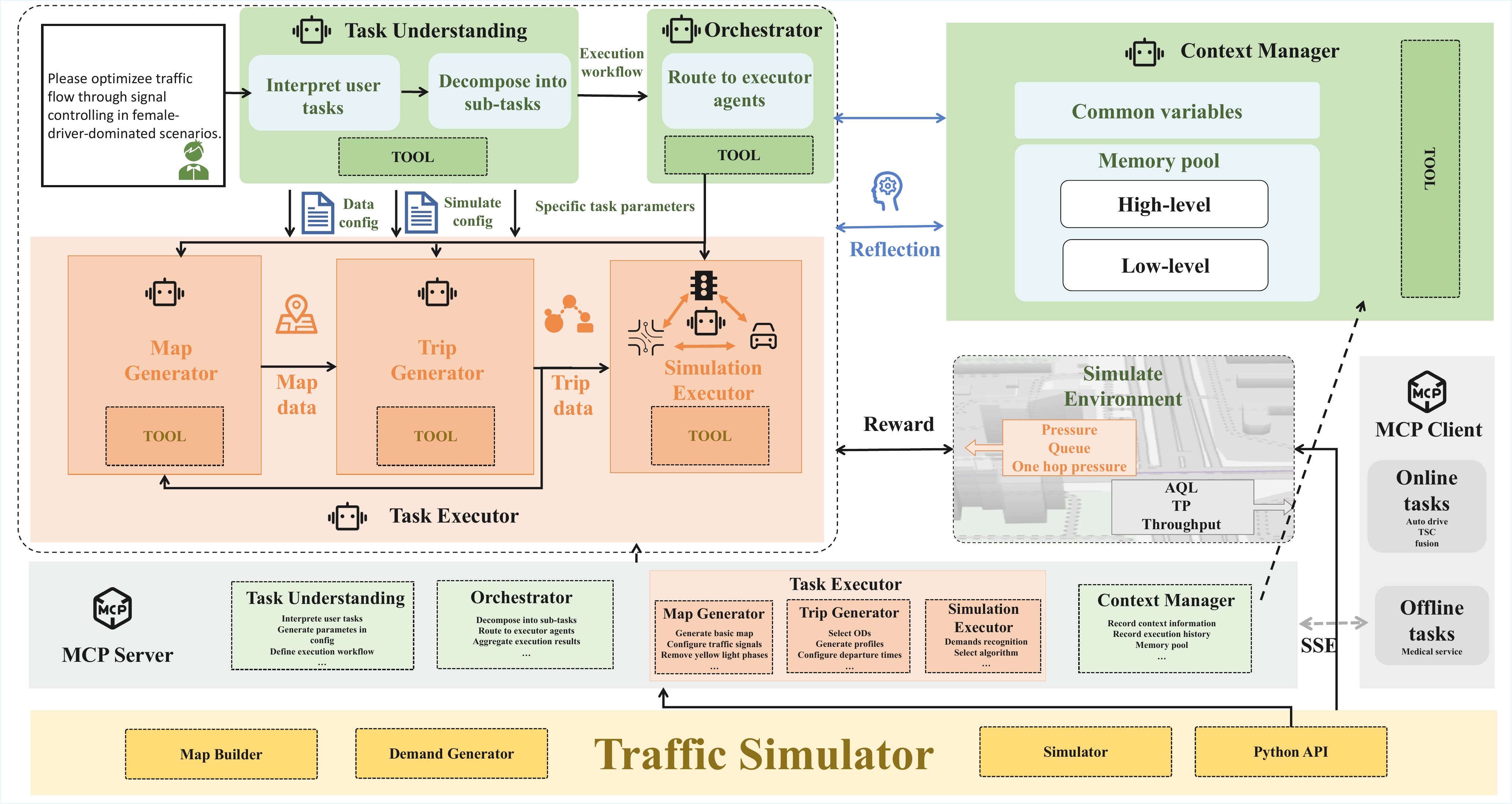}
    \caption{The whole framework of \textit{\systemname~}.}
    \label{fig:framework}
\end{figure*}

The main contributions of this paper are summarized as follows:
\begin{itemize}[leftmargin=1.5em,itemsep=0pt,parsep=0.2em,topsep=0.0em,partopsep=0.0em]
\item To the best of our knowledge, we are the first to propose an autonomous traffic simulation agent capable of both automatic optimization and scenario generalization.
\item We introduce a set of Abstracted API Functions, decoupled from the underlying traffic simulation platform and exposed via the MCP interface, establishing a foundational infrastructure for general-purpose autonomous traffic simulation.
\item We design a novel Task Understanding and Autonomous Planning module that supports the interpretation of ambiguous natural language instructions and enables multi-scenario generalization, thus allowing the system to adapt to diverse environments.
\item We propose an Embedded Optimization Module featuring a two-layer optimization system: one based on fully autonomous LLM-driven optimization, and another combining the LLM with classical algorithms. This enables self-optimization of the simulation and performance improvement.
\item Extensive experiments demonstrate the generalization and stability of our proposed framework across various simulation scenarios. Furthermore, its automatic optimization capability enables automated, iterative performance improvement tailored to specific task requirements.
\end{itemize}

\section{Methods} \label{sec:methods}
\systemname~is designed as a collaborative multi-agent framework that bridges natural language instructions and traffic simulation execution. As shown in Figure\ref{fig:framework}, our framework consists of four core modules working together: Task Understanding Module, Orchestrator Module, Task Executor Module, and Context Manager Module. Unlike open-ended agent frameworks like OpenManus that lack task-specific structure, \systemname~incorporates domain knowledge of transportation experiments through specialized module design, ensuring both flexibility and execution reliability.
\subsection{Overall Framework}
LLM agents act as experiment planners and optimizers. According to the general implementation of transportation experiments, we design a task understanding module, three task execution modules—including a map generator module, a trip generator module, and a simulation executor module—each managed by an LLM agent. To assist each agent in better understanding the specific implementation plan and enabling efficient collaboration among agents, we design an Orchestrator module that serves as a global planner and a Context Manager that functions as a shared memory pool.
The workflow begins with the Task Understanding Module interpreting user instructions. The Orchestrator then decomposes the task into subtasks and routes them to appropriate executors. Each executor module organizes experiments automatically and invokes necessary tools via the MCP server. Throughout this process, the Context Manager maintains execution history and agent memories, enabling reflection and learning. The server, built on the MCP-agent framework, dynamically registers and runs only the necessary tools—such as MOSS utilities and other auxiliary utilities—via a lightweight JSON-RPC interface.
\subsubsection{MCP client}

\textbf{Task Understanding Module} The module understands the user's instructions, then recognizes key parameters for each executor module; detailed parameters are provided in Table\ref{tab:system_architecture}. The module also performs global planning, assigning the execution order of each module according to the user's instructions.
For example, the instruction:
"compare the TSC experimental results of Yizhuang during morning peak hour and Shanghai at midnight"
corresponds to the following execution order:
map generator → trip generator → simulation executor → map generator → trip generator → simulation executor.

\textbf{Task Executor Module} The module consists of three sub-modules.

(1)Map Generator: Generate traffic network maps according to the parameter region. First, use the geocoding service of OSM (OpenStreetMap) to obtain the coordinates of the four corners of the region boundary. Then, fetch POI, AOI, street, lane, and junction data within the boundary from OSM in GeoJSON format. Finally, use GeoBuilder to convert the GeoJSON to Protocol Buffer (Protobuf) format, while also performing lane reconstruction and traffic light generation within intersections.

(2)Trip Generator: Generate trip files for vehicles and persons for simulation. First, generate user profiles and departure time curves in parallel with the distribution described in the user instructions. Then, generate the OD matrix (Origin-Destination matrix) for persons and vehicles (drivers) in the target region. Finally, set personalized driving patterns for each driver and use routing algorithms to generate travel paths based on the OD matrix. It also supports post-processing map data if required for specific tasks. For example, removing the yellow light phase from the map for TSC (Traffic Signal Control) tasks.

(3)Simulation Executor: Execute simulation tasks and choose the optimal execution strategy according to the user instructions. We categorize tasks into online simulation and offline simulation. Offline simulation involves modifying static elements, such as AOI, street, lane, and junction data. Online simulation involves modifying dynamic elements by analyzing real-time information collected during the simulation, such as adjusting vehicle driving states based on road conditions and changing traffic light phases at intersections in real time. Online simulation supports both traditional RL algorithms and LLM agent-controlled methods to collect environmental information, calculate current rewards, and determine update strategies for future states of dynamic elements.

\textbf{Orchestrator Module}
This module functions as an agent router, decomposing complex tasks into sub-tasks and routing each sub-task to the relevant module's agent. It supports multi-agent parallel task execution, monitors the execution process, and aggregates the execution results from each module.

\textbf{Context Manager Module}
This module consists of a global context manager and multiple individual memory managers. The global context manager oversees the execution results of each module and shares key parameter information between modules. Once the task understanding module receives the user’s instructions, the context manager creates a new session to record the context history of each task executor module, including tool parameters, tool execution order, and error messages. Task executor modules can perform reflection based on this history to enhance task execution success rates. The memory managers record the decision history of each agent in the simulation executor module, adaptively updating as simulation time progresses, enabling agents to recognize and learn from the highest-reward decisions.
\subsubsection{MCP Server}
\textbf{Tools} The tool set is mainly organized into four groups, corresponding to the task understanding module and the three task execution modules. Each group remains inactive until the corresponding module is executed. There are also some auxiliary tools in Table\ref{tab:system_architecture} to facilitate a higher task execution success rate.
\begin{figure}
    \centering
    \includegraphics[width=\linewidth]{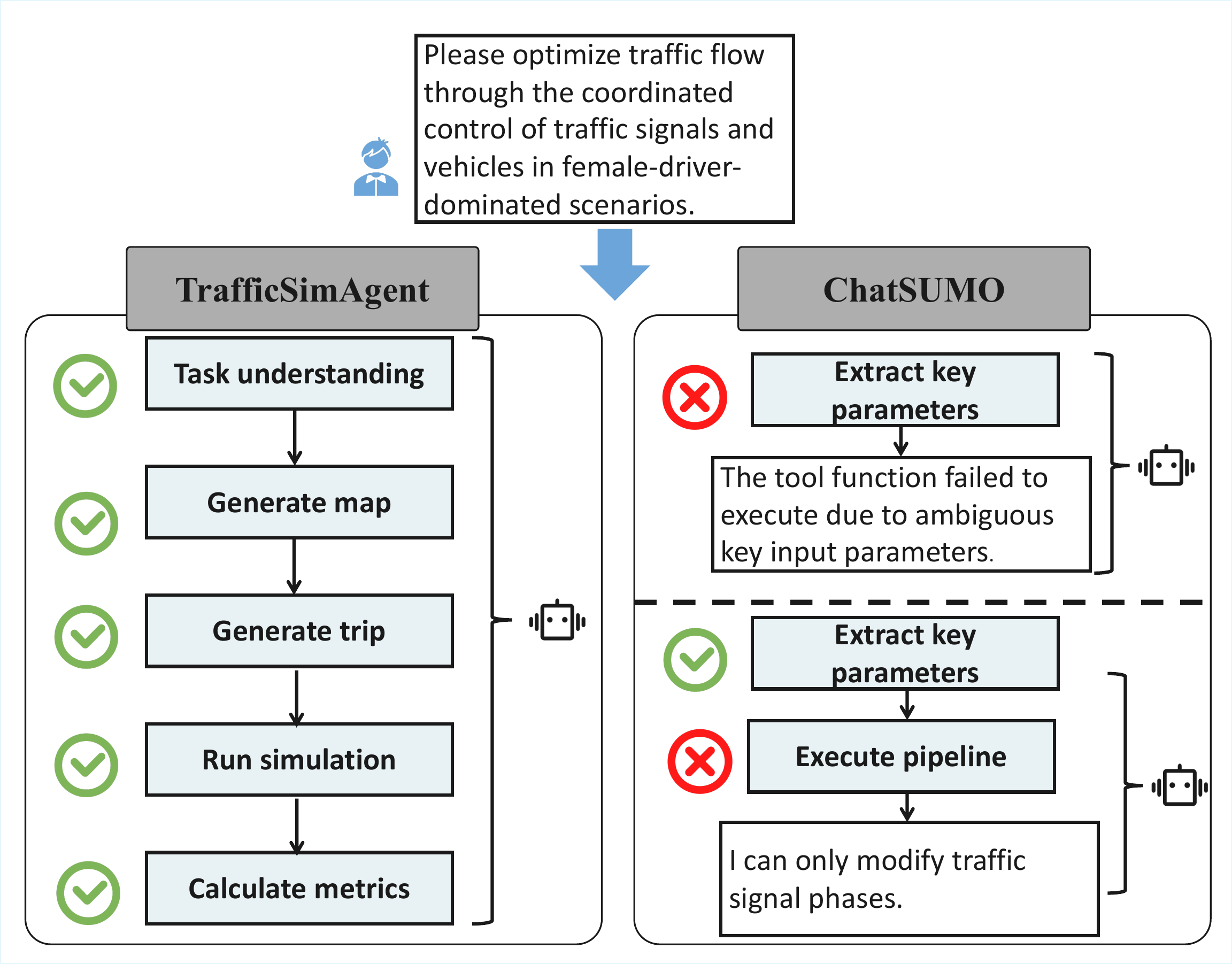}
    \caption{Case study example}
    \label{fig:case}
\end{figure}

\subsection{Design for Enhanced Generality}
The architecture of \systemname~ is specifically designed to achieve superior generality across diverse simulation scenarios, overcoming the limitations of previous systems that rely on predefined workflows and limited task support. This enhanced generality is realized through three key design innovations: robust instruction understanding, dynamic task planning, and flexible element-agent embodiment.

\subsubsection{Robust Instruction Understanding through Semantic Comprehension}
Unlike pattern-based approaches such as ChatSUMO that extract predetermined parameters through template matching, \systemname~'s Task Understanding Module employs deep semantic comprehension to handle ambiguous, incomplete, or context-dependent instructions. The module utilizes LLM-powered analysis to infer implicit parameters and contextual relationships, then structures this understanding into standardized configuration files for downstream processing.
This approach significantly expands the range of acceptable instructions beyond predefined templates, enabling natural and flexible user interaction while maintaining structured data flow between modules.

\subsubsection{Dynamic Task Planning through Hierarchical Decomposition}
Traditional systems like ChatSUMO employ fixed execution workflows that limit their applicability to predetermined scenarios. \systemname~ overcomes this limitation through its Orchestrator Module, which dynamically plans and adapts execution sequences based on semantic understanding of each unique instruction. We show a case in Figure\ref{fig:case}.

The dynamic planning process enables handling of complex, multi-faceted instructions that would require manual workflow specification in previous systems. For instance, the comparative instruction "analyze traffic patterns in Shanghai between morning and evening peaks, then optimize signal timing for the worst-performing period" triggers the following adaptive workflow:
(1)Map generation for Shanghai region.
(2)Trip generation with morning peak profile parameters.
(3)Simulation execution and metric collection.
(4)Trip regeneration with evening peak profile parameters.
(5)Simulation execution and metric collection.
(6)Comparative analysis and bottleneck identification.
(7)Targeted optimization of identified critical intersections.
(8)Validation simulation and results reporting.
This flexible planning capability allows \systemname~ to accommodate diverse instruction types without requiring predefined workflow templates, significantly expanding its applicability across different experimental designs.
\subsubsection{Diverse Task Support through Element-Agent Embodiment}
The framework's generality is further enhanced by its foundational principle of element-agent embodiment, where each fundamental traffic element (e.g., an intersection, a vehicle) is represented by an LLM agent that makes autonomous decisions in real-time based on environmental information. This architecture supports a wide range of simulation tasks through emergent agent behaviors rather than hardcoded scenario implementations.
For example, the fusion task requires coordinated decision-making between agents role-playing traffic signals and those role-playing vehicles. Each agent optimizes its behavior while considering the others' states and intentions, without the need for pre-programmed coordination logic. 
\subsection{Design for Autonomous Optimization}
\systemname~'s autonomous optimization capability is achieved by employing LLM agents to act as fundamental traffic elements and make coordinated decisions based on real-time states, moving beyond merely functioning as an on/off switch for the simulation process. This optimization operates at two levels: high-level optimization strategy selection and low-level real-time collaborative optimization. The Context Manager Module enables agents to reflect on their execution records and rewards from environment, continuously improving success rates.
\subsubsection{High-Level Optimization Strategy Selection}
The Orchestrator Module is able to automatically select the most suitable optimization approach based on user instructions and task characteristics. For tasks requiring precise numerical optimization (e.g., "minimize average travel time"), it may invoke traditional optimization algorithms such as MaxPressure or reinforcement learning-based methods. For tasks demanding commonsense reasoning and complex trade-offs (e.g., "reduce congestion while maintaining safety"), it employs LLM-based optimization strategies. This adaptive selection mechanism ensures optimal performance across diverse scenarios.
\subsubsection{Low-Level Real-Time Collaborative Optimization}
During simulation execution, low-level agents achieve real-time closed-loop optimization of traffic conditions through perception-decision-action cycles. Each agent (e.g., traffic signal or vehicle) makes decisions considering not only its own state but also the cooperative status of other elements and real-time traffic conditions information. For instance, a traffic signal agent adjusts phase timing by comprehensively analyzing current queue lengths, pressure in the neighbourhood, the intentions of approaching vehicles, and overall regional traffic density. Similarly, a vehicle agent selects optimal acceleration and speed while considering signal states ahead and the cooperative behavior of surrounding vehicles. This multi-element collaborative decision-making mechanism overcomes the limitations of traditional algorithms (e.g., MaxPressure, MPLight) that utilize only localized neighborhood information, and existing LLM-based methods (e.g., LLMLight) that inadequately consider the status of other elements

\begin{table*}
\centering
\caption{Modular architecture and functional components of \systemname.}
\setlength{\tabcolsep}{1.5mm} %
\small %
\renewcommand{\arraystretch}{1.2} %
\resizebox{1\textwidth}{!}{
\begin{tabular}{c c c c}
\toprule
\textbf{module name} & \textbf{submodule name} & \textbf{tools} & \textbf{parameters} \\
\midrule
Task Understanding & -- & \makecell[c]{analyze-requirement, extract-key-parameter,\\ validate-parameters} & natural\_language\_input \\
Orchestrator & -- & agent\_router & -- \\
\multirow{3}{*}{Task Executor} & Map Generator & \makecell[c]{generate-basic-map, configure-traffic-signals,\\ preprocess-map-for-tsc} & \makecell[c]{region name,\\ green\_time, yellow\_time} \\
& Trip Generator & \makecell[c]{select-origins-destinations, generate-profiles,\\ configure-departure-times, generate-persons-vehicles,\\ configure-personalized-driving} &\makecell[c]{boundary\_coordinates, age\_ranges,\\ cons\_ranges, gender\_distribution,\\ education\_distribution, start\_step, duration\_step,\\ persons\_num, vehicles\_num} \\
& Simulation Executor & \makecell[c]{demand-recognition, select-algorithm,\\ execute-scenario, monitor-simulation-progress,\\ extract-simulation-metrics} & \makecell[c]{algorithm, scenario\_name, reward\_type,\\ llm\_control\_interval, start\_step, duration\_step} \\
\multirow{2}{*}{Context Manager} & Common Variables & \makecell[c]{create\_session, get\_session, export\_session, import\_session,\\ record\_tool\_call, get\_tool\_call\_history,\\ get\_agent\_state, update\_agent\_state} & session\_id \\
& Memory Pool & \makecell[c]{get\_agent\_background, record\_decision,\\ add\_background\_knowledge, search\_conversation,\\ get\_conversation\_summary, clear\_memory, health\_check} & \makecell[c]{session\_id, max\_conversation\_length,\\ max\_session\_memory, memory\_retention\_days,\\ auto\_summarize\_interval, background\_info\_length,\\ max\_context\_variables, context\_summary\_length} \\
\bottomrule
\end{tabular}}
\label{tab:system_architecture}
\end{table*}

\section{Experiments} \label{sec:exp}
\systemname~utilizes MOSS as the simulator with an MCP-agent. Users input natural language instructions at the client side, and related tools are subsequently invoked at the server side. Variables parsed from these instructions and intermediate variables (for example, data paths and the boundary coordinates of the target region) are stored in the context manager. Instead of relying on a predefined workflow and setup, we employ an Orchester module to plan the execution sequence and assign reasonable values to parameters in the configuration file, guided by the natural language instructions. We also enable LLM agents to optimize the experimental results during the simulation process, rather than merely functioning as an on/off switch. Therefore, we primarily aim to demonstrate the versatility and autonomous optimization capability of \systemname.
\subsection{Experimental Settings}
\subsubsection{Baseline Models and Methods}

To comprehensively evaluate the performance of \systemname, we compare it against a range of baseline approaches, categorized into two groups for the main experiment and the optimization experiment, respectively.

\textbf{Baselines for Main Experiment} We select baselines that represent different paradigms in task automation and language-driven simulation:
\begin{itemize}
\item Domain-Specific Agent Frameworks: ChatSUMO~\cite{li2024chatsumo}, a specialized framework that integrates LLMs with the SUMO simulator for traffic simulation tasks.
\item General-Purpose Agent Frameworks: OpenManus~\cite{openmanus2025} and MetaGPT~\cite{hong2024metagpt}, which are state-of-the-art, multi-agent frameworks designed for general software task execution but not specifically tailored for traffic simulation.
\item General Large Language Models (LLMs): GPT-5 and Gemini 2.5 Pro~\cite{comanici2025gemini}. These powerful models are prompted to perform the simulation tasks in a zero-shot or few-shot manner, representing the capability of raw LLMs without specialized frameworks.
\end{itemize}
\textbf{Baselines for Optimization Experiment} To specifically assess the optimization capability of our low-level agents, we compare against established Traffic Signal Control (TSC) algorithms:
\begin{itemize}
\item MaxPressure~\cite{liu2024max}: A classic, model-free TSC algorithm that controls signals based on the real-time pressure (queue length) of incoming and outgoing lanes.
\item MPLight~\cite{chen2020toward}: A deep reinforcement learning-based method that utilizes the MaxPressure principle as part of its reward function for adaptive signal control.
\item LLMLight~\cite{lai2025llmlight}: A state-of-the-art LLM-based TSC method that leverages large language models for decision-making at intersections.
\end{itemize}
\subsubsection{Dataset Construction}

All experimental data were generated on-demand by our \systemname~via the MOSS simulation environment, including a map network and travel demand data. The data generation was directly guided by the user's natural language instructions, ensuring each scenario accurately reflects the specified conditions.

\textbf{Map Network Dataset} comprises the fundamental topological and geographical elements required for traffic simulation, which is generated by the Map Generator in executor module through the following procedure:

(1)Geocoding: Given a region name (e.g., "Yizhuang, Beijing"), we utilize the OpenStreetMap (OSM) geocoding service to obtain its precise geographical boundaries (i.e., the four-corner coordinates).

(2)Data Fetching \& Conversion: The bounding box is used to fetch raw map data from OSM, which is then converted into a simulation-ready network using the MOSS simulator's utilities. The final map dataset comprises several core elements: lanes, roads, junctions, and AOIs (Areas of Interest).

\textbf{Travel Demand Dataset} defines the movement of individuals and vehicles, which is generated by the Trip Generator in executor module through the following procedure:

(1)OD Matrix Generation: An Origin-Destination (OD) matrix is first generated based on an analysis of map dataset generated by Map Generator.

(2)Individual Demand Synthesis: The aggregate OD matrix is then disaggregated into individual travel demands. This critical step is guided by the natural language instruction and involves intelligently determining several key parameters. To enhance the robustness of this process, Context Manager Module enables reflective validation, ensuring the selected parameters and scenario configurations are contextually appropriate and consistent with the user's intent. Key parameters are as follows: 
\begin{itemize}
\item Population Size: The total number of users to simulate.
\item Origin-Destination Selection: Choosing specific start and end points (AOIs) that reflect the instructed scenario.
\item Travel Mode: Assigning modes of transport (e.g., car, taxi, walk).
\item User Profiles: Generating demographic profiles for each individual, including age, gender, education, and consumption level.
\item Departure Time Distribution: Configuring a time distribution curve for trip departures (e.g., a peak-shaped curve for "rush hour").
\item Driving Parameters: Configuring either personalized or unified driving behavior parameters (e.g., usual acceleration, average speed) for each vehicle.
\end{itemize}
(3) Simulation Integration: The generated individual trips are finally integrated with the road network for execution in the simulator.
The key parameters used in the travel demand generation are summarized in Table~\ref{tab:system_architecture}. 
\subsubsection{Metrics}
To comprehensively evaluate the performance of all methods, we employ a set of metrics from traffic simulation, categorized into efficiency and environmental impact dimensions. Efficiency is measured by traffic volume (TV), cumulative throughput, travel time (ETA, ATT-finished), queue length (AQL), and ETA reduction ($\Delta$ETA); environmental impact is gauged by total and average carbon emissions (Carbon, ACE). Formal definitions are provided in Appendix~~\ref{sec:appendix}.

\vspace{-5pt}
\subsection{Generalization}
\begin{table*}[!htpb]
  \centering
  \small
  \caption{Performance comparison of simulation methods. Best and second-best results are highlighted in \textbf{bold} and \underline{underline}, respectively.}
  \label{tab:comparison-simulation-results}
  \resizebox{\linewidth}{!}{%
  \begin{tabular}{l|ccccccccccccc}
    \toprule
    \multirow{3}{*}{Model} & \multicolumn{3}{c}{auto drive} & \multicolumn{3}{c}{fusion} & \multicolumn{3}{c}{tsc} & \multicolumn{3}{c}{medical service} & \multirow{3}{*}{MRR$\uparrow$} \\ 
    \cmidrule(lr){2-4} \cmidrule(lr){5-7} \cmidrule(lr){8-10} \cmidrule(lr){11-13}
    & TV$\uparrow$ & AQL$\downarrow$ & TP$\uparrow$ & AQL$\downarrow$ & \begin{tabular}{@{}c@{}}ACE$\downarrow$\end{tabular} & TP$\uparrow$ & AQL$\downarrow$ & \begin{tabular}{@{}c@{}}ACE$\downarrow$\end{tabular} & TP$\uparrow$ & serve\_rate$\uparrow$ & ATT$\downarrow$ & \begin{tabular}{@{}c@{}}score$\uparrow$\end{tabular} & \\ 
    \midrule
    ChatSUMO & - & - & - & - & - & - & \textbf{18.91} & 0.59 & 673 & - & - & - & 0.246 \\ 
    gpt-5 & \underline{408} & 17.04 & 1116 & \underline{18.63} & 1.44 & 1136 & 30.43 & 0.30 & 1253 & 98\% & 3064.29 & -0.233 & 0.336 \\ 
    gemini2.5-pro & 401 & \underline{17.11} & 1121 & 25.95 & 0.40 & 1290 & 22.31 & 0.69 & 1255 & 98.5\% & 1472.81 & 0.248 & 0.285 \\ 
    MetaGPT & 94 & 18.29 & 677 & \textbf{18.34} & 0.75 & 677 & \underline{19.86} & 0.59 & 673 & 91.0\% & 3665.58 & -0.463 & 0.321 \\ 
    OpenManus & 125 & 74.90 & \textbf{2113} & 74.90 & \underline{0.32} & \textbf{2113} & 74.95 & \textbf{0.25} & \textbf{2110} & 99.5\% & 1283.82 & 0.311 & \underline{0.651}\\ 
    \systemname~& \textbf{420} & \textbf{16.81} & \underline{1308} & 24.34 & \underline{0.29} & \underline{1417} & 29.10 & \underline{0.29} & \underline{1337} & \textbf{99.5\%} & \textbf{1283.82} & \textbf{0.311} & \textbf{0.715} \\ 
    \bottomrule
  \end{tabular}%
  }
\end{table*}

\begin{figure}[h]
    \centering
    \includegraphics[width=\columnwidth]{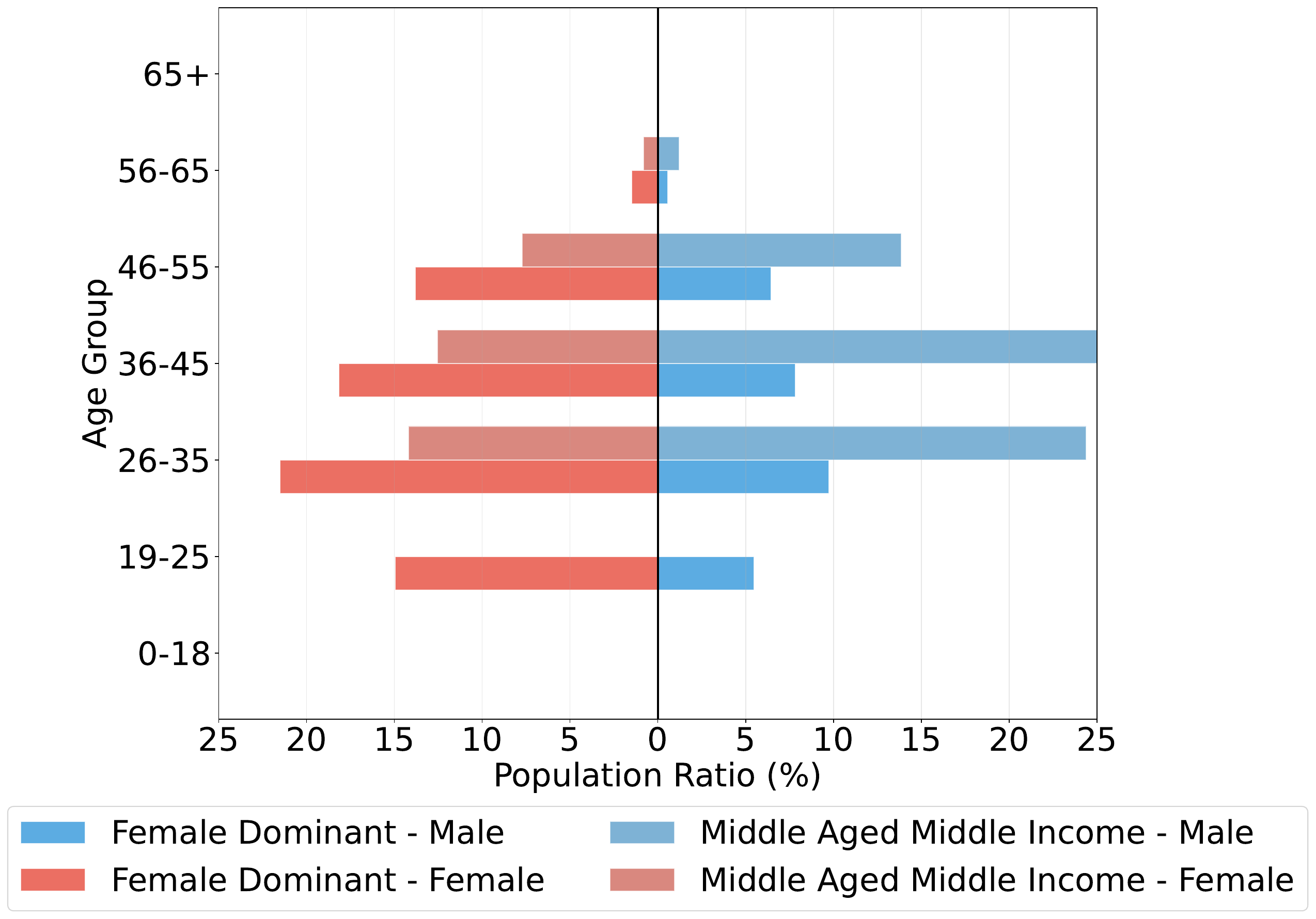}
    \caption{Gender and age distribution of user data generated by \systemname~for different user groups.}
    \label{fig:profiles-age}
\end{figure}

\begin{figure}[h]
    \centering
    \includegraphics[width=1\columnwidth]{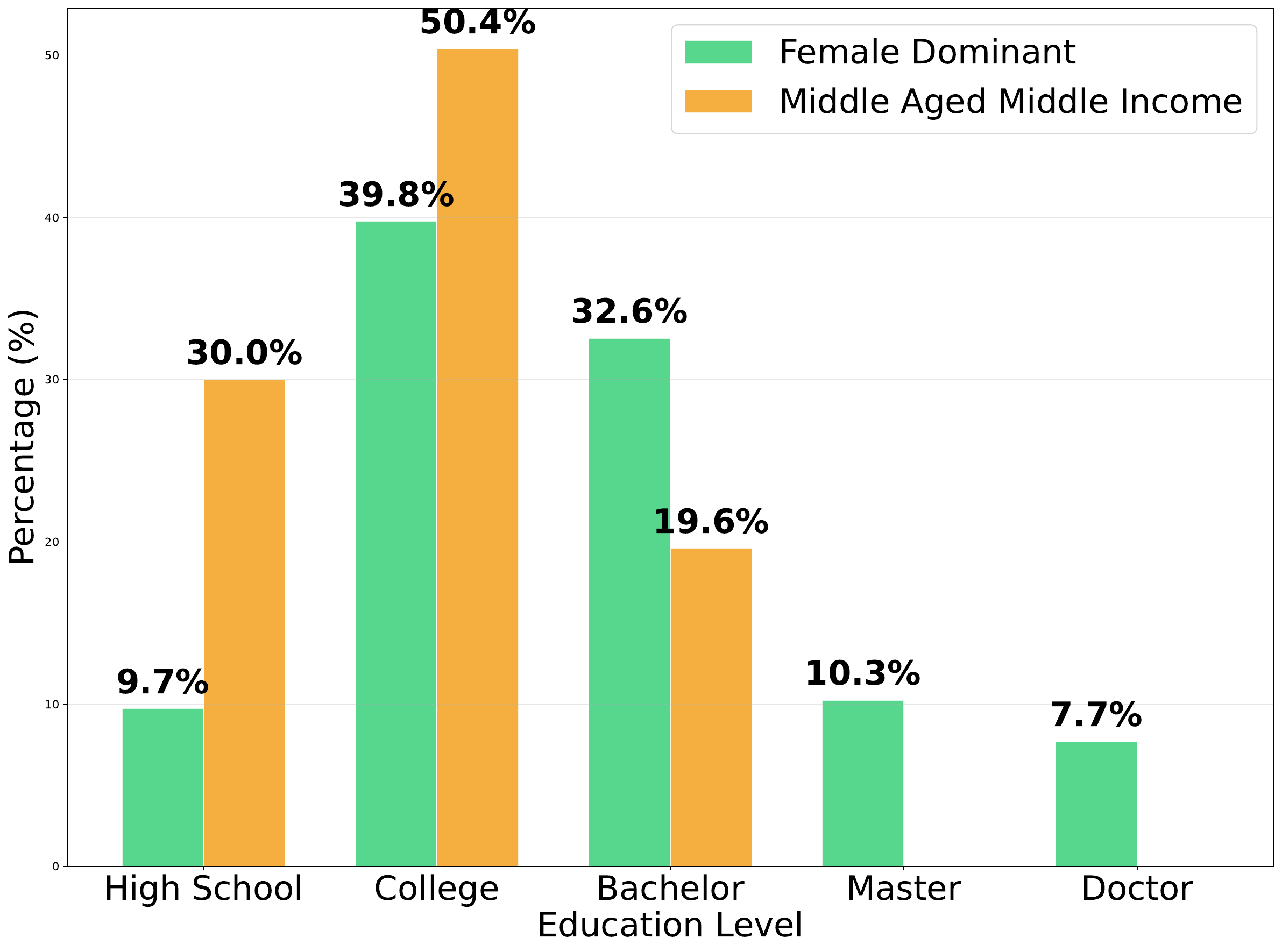}
    \caption{Education distribution of user data generated by \systemname~for different user groups.}
    \label{fig:profiles-edu}
\end{figure}

To demonstrate the ability of \systemname~to handle various types of natural language instructions, we designed a natural language instruction benchmark covering diverse scenarios across different simulation tasks, categorized into "specified" and "ambiguous" instructions. The model was tasked with two ambiguous instruction to simulate driving behaviors of two distinct demographic groups: middle-aged, middle-income drivers with median education levels(Middle Aged Middle Income), and female-dominated driving populations(Female Dominant). The generated results are presented in Figure~\ref{fig:profiles-age} and Fig~\ref{fig:profiles-edu}, showing distributions that align broadly with the specified instructions, illustrating that it genuinely understands ambiguous instructions instead of mechanically parsing parameters from them.

Furthermore, we sought to evaluate the general capability of \systemname~on both online and offline simulation tasks. We selected medical service selection as the offline simulation task, and auto\_drive, TSC (Traffic Signal Control), and fusion as the online simulation tasks. We compared the results of \systemname~against a general model, a general agent framework, and a domain-specific agent framework. Table~\ref{tab:comparison-simulation-results} presents the comparison results.

The results indicate that ChatSUMO, the domain-specific model, can only handle the TSC task. GPT-5 and Gemini-pro, the general models, show limited performance in either congestion mitigation (high AQL) or economic and environmental efficiency (high carbon emission). MetaGPT, as a general agentic framework, adopts an overly conservative strategy (low TP, low AQL), whereas OpenManusor adopts the opposite approach (high TP, high AQL). In contrast,\systemname~strikes a balance among TP, AQL, and carbon emission, enabling all vehicles to complete their trips as quickly and smoothly as possible without excessive congestion, thereby avoiding carbon emissions caused by inefficient acceleration near the speed limit.

\vspace{-5pt}
\subsection{Autonomous Optimization}
\begin{table*}[!htpb]
  \centering
  \small
  \caption{Performance comparison of traffic condition optimization methods. Best and second-best results are highlighted in \textbf{bold} and \underline{underline}, respectively.}
  \label{tab:opt-methods-results}
  \resizebox{0.85\linewidth}{!}{%
  \begin{tabular}{llcccccccccc}
    \toprule
     & Opt Methods & \begin{tabular}{@{}c@{}}ATT-f$\downarrow$\end{tabular} & \begin{tabular}{@{}c@{}}TV$\uparrow$\end{tabular} & \begin{tabular}{@{}c@{}}AQL$\downarrow$\end{tabular} & \begin{tabular}{@{}c@{}}$\Delta\text{ETA}$$\uparrow$\end{tabular} & \begin{tabular}{@{}c@{}}ACE$\downarrow$\end{tabular} & \begin{tabular}{@{}c@{}}ETA$\downarrow$\end{tabular} & \begin{tabular}{@{}c@{}}TCE$\downarrow$\end{tabular} & \begin{tabular}{@{}c@{}}TP$\uparrow$\end{tabular} & \begin{tabular}{@{}c@{}}\textbf{MRR}$\uparrow$\end{tabular} \\
    \midrule
    \multirow{3}{*}{\makecell{Others\\(TSC)}} & MPLight & 165.15 & 143 & 17.44 & -161.98 & \underline{0.84} & 15487651.37 & \underline{440.92} & 526 & 0.360 \\
                            & MaxPressure & 200.27 & 215 & 18.9142 & -164.79 & 0.87 & \textbf{16253681.12} & 476.05 & 551 & 0.257 \\
                           & LLMLight & 312.92 & \textbf{1027} & \textbf{13.486} & -162.53 & 1.44 & \underline{17836227.97} & 853.84 & 595 & 0.405 \\
    \midrule
    \multirow{3}{*}{\makecell{Ours \\(CollabOpt)}} & auto-drive & 355.09 & \underline{420} & \underline{16.81} & -142.35 & 1.22 & 33166900.16 & 1355.3 & 1308 & 0.290 \\
                               & fusion & \underline{134.68} & 120 & 24.34 & \textbf{-125.73} & 0.29 & 39663544 & \textbf{420.67} & \textbf{1417} & \textbf{0.564} \\
                               & TSC & \textbf{125.42} & 105 & 29.10 & \underline{-139.53} & \textbf{0.29} & 39634600.37 & 387.76 & \underline{1337} & \underline{0.557} \\
    \midrule
    no-control & -- & 354.55 & 420 & 25.07 & -156.75 & 1.13 & 32943916.96 & 1249.22 & 1107 & 0.244 \\
    \bottomrule
  \end{tabular}%
  }
\end{table*}

We evaluated the traffic optimization results by comparing \systemname~with traditional TSC algorithms (MaxPressure, MPLight) and LLMLight, a state-of-the-art LLM-based TSC algorithm. Experiments were performed using the aforementioned dataset. The comparison results are shown in Table~\ref{tab:opt-methods-results}. \systemname, which employs LLM agents for macro-level traffic condition optimization by controlling individual traffic elements in consideration of current traffic conditions and the status of other elements, proves to be the optimal approach. Specifically, the fusion algorithm performs best by enabling cooperative control between traffic lights and vehicles. LLMLight fails to adequately consider the information from other elements, while traditional algorithms only utilize neighborhood information, leading to a suboptimal overall optimization outcome.

\subsection{Analysis on Agent Reasoning and Model Scaling}

\begin{figure}[h]
    \centering
    \includegraphics[width=1\columnwidth]{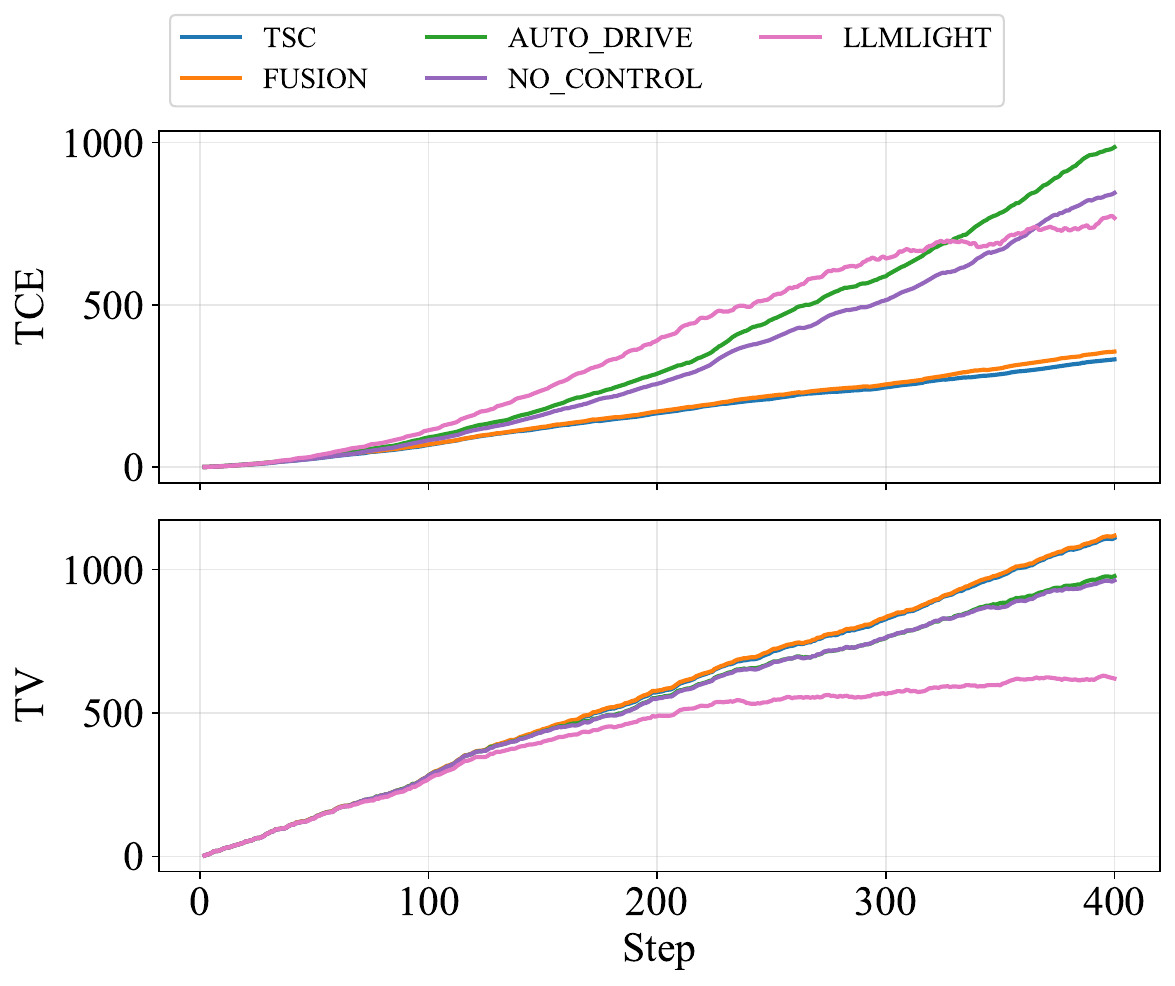}
    \caption{Metrics variation with simulation steps of different optimization methods.}
    \label{fig:comparison}
\end{figure}

To analyze the advantages of our framework over traditional LLM-based TSC implementations like LLMLight, which typically provides junction phase information to traffic signal agents at each step, we compare the evolution of traffic throughput and carbon emission across simulation steps, as shown in Figure~\ref{fig:comparison}.
The experimental results reveal that our method's superiority becomes increasingly pronounced as simulation progresses. This emergent advantage stems from a fundamental difference in decision-making paradigms. While conventional approaches like LLMLight rely on fine-grained phase analysis (providing per-phase queue lengths for each junction), our low-level agents operate under a reinforcement learning-inspired philosophy that maximizes a composite reward function incorporating pressure differences between incoming and outgoing lanes, total queued vehicles, and neighboring junction influences.

This reward-driven approach creates a critical shift from short-term reactive decisions to long-term strategic optimization. The phase-specific information used by traditional methods often leads to greedy decisions that minimize immediate queue lengths but disrupt the balance between intersection pressure and overall traffic throughput. In contrast, our agents leverage historical trend analysis through the scratchpad mechanism (e.g., strategy: try-better-action) to learn from long-term outcomes rather than reacting to instantaneous snapshots.

Our goal-oriented reward signal combined with memory-enabled strategic decision-making drives more effective long-term optimization. Besides, Historical State maintains strategy effectiveness through action-reward memory; Other Junctions enables coordinated multi-agent optimization; and the Scratchpad provides essential trend analysis for adaptive decision-making.
In conclusion, our framework's growing advantage over traditional phase-analysis approaches underscores the effectiveness of replacing detailed perceptual data with goal-driven, history-aware decision processes for complex dynamic optimization in traffic signal control.

Our experiments with varying model sizes in Table~\ref{tab:models-results} demonstrate a clear positive correlation between parameter count and performance. The Qwen3-235B model achieves overall superior results, while smaller models show progressively reduced capability. This confirms that \systemname~'s performance scales with the underlying model's capacity.

\begin{table*}[!htpb]
  \centering
  \small
  \caption{Performance comparison of different models. Best and second-best results are highlighted in \textbf{bold} and \underline{underline}, respectively.}
  \label{tab:models-results}
  \resizebox{0.85\linewidth}{!}{%
  \begin{tabular}{lccccccccccc}
    \toprule
    Model & \begin{tabular}{@{}c@{}}ATT-f$\downarrow$\end{tabular} & \begin{tabular}{@{}c@{}}TV$\uparrow$\end{tabular} & \begin{tabular}{@{}c@{}}AQL$\downarrow$\end{tabular} & \begin{tabular}{@{}c@{}}$\Delta\text{ETA}$$\uparrow$\end{tabular} & \begin{tabular}{@{}c@{}}ACE$\downarrow$\end{tabular} & \begin{tabular}{@{}c@{}}ETA$\downarrow$\end{tabular} & \begin{tabular}{@{}c@{}}TCE$\downarrow$\end{tabular} & \begin{tabular}{@{}c@{}}TP$\uparrow$\end{tabular} & \begin{tabular}{@{}c@{}}SR$\uparrow$\end{tabular} & \begin{tabular}{@{}c@{}}MRR$\uparrow$\end{tabular} \\
    \midrule
    mistral7b-v3 & 365.25 & \textbf{215} & \textbf{12.25} & \underline{-146.26} & 2.27 & 37595153.88 & 2869.91 & \underline{1266} & 35\% & 0.344 \\
    llama3.1-8b & \underline{352.85} & \underline{316} & \underline{14.74} & -152.77 & 1.90 & 35256368.16 & 2248.59 & 1186 & 47\% & 0.287 \\
    qwen3-14b & 355.28 & 391 & 16.51 & -155.29 & \textbf{1.11} & 
    33556686.51 & \textbf{1256.77} & 1128 & 49\% & \underline{0.424} \\
    deepseek-v3 & \textbf{350.02} & 403 & 16.78 & -156.00 & 1.32 & \underline{33383975.97} & 1480.04 & 1122 & 62\% & 0.393 \\
    llama3-70b & 370.33 & 335 & 15.56 & -150.59 & 1.62 & 35116956.06 & 1918.44 & 1181 & \underline{65\%} & 0.287 \\
    qwen3-235b & 355.09 & 420 & 16.81 & \textbf{-142.35} & \underline{1.22} & \textbf{33166900.16} & \underline{1355.3} & \textbf{1308} & \textbf{78\%} & \textbf{0.611} \\
    \bottomrule
  \end{tabular}%
  }
\end{table*}

\section{Related Work} \label{sec:related}
\subsection{Traffic Simulation}
Traffic simulation has long been established as a cornerstone of urban computing and intelligent transportation systems research\cite{tao2023flowsim, van2016coordinated,yazdani2023intelligent}. High-fidelity, open-source simulators such as SUMO~\cite{behrisch2011sumo}, MOSS~\cite{zhang2024moss}, MATSim~\cite{w2016multi} and CityFlow~\cite{zhang2019cityflow} provide researchers with powerful platforms for modeling complex traffic dynamics, evaluating traffic management policies, and testing autonomous driving algorithms. There are also traffic simulator for auto-drive, such as LCSim~\cite{zhang2024lcsim} and CARLA~\cite{dosovitskiy2017carla}. These tools offer extensive libraries for network modeling, vehicle movement, and traffic light control.
However, conducting simulation experiment remains a non-trivial task that often poses significant challenges for users. For example, many autonomous driving-related studies based on SUMO are designed to perform each simulation step manually, such as SMDT~\cite{wang2024smart}, BlaFT~\cite{park2023blame}, HPTSim~\cite{shi2024hptsim}, DLIO~\cite{wang2024iterative}, ASC-HBMP~\cite{zong2023human}, and RoboCar~\cite{testouri2025robocar}. Moreover, the very process they manually execute is inherently complex, typically involving multiple intricate steps: acquiring and preprocessing map data (e.g., from OpenStreetMap), defining traffic demand (e.g., generating trip files and routes), configuring simulation parameters, and finally, executing and monitoring the simulation. This workflow requires considerable domain expertise and is largely manual, making it time-consuming, error-prone, and difficult to reproduce. Furthermore, while these simulators are capable of modeling various scenarios, they offer limited native support for automatic optimization. 

\subsection{LLM-Driven Simulation Automation}
Recent works have begun leveraging large-language models (LLMs) within agentic frameworks to automate complex tasks through natural language instructions, demonstrating significant potential in lowering the barrier to entry for non-expert users. The emergence of autonomous AI agents has showcased LLMs' capability in task decomposition and sequential execution, while specialized systems such as ChemCrow~\cite{bran2023chemcrow} have demonstrated expert-level automation in scientific domains like organic synthesis and drug discovery. In cybersecurity, AutoPentest~\cite{henke2025autopentest} exemplifies how LLM agents can autonomously conduct penetration testing and vulnerability assessment. In trajectory modeling, TrajAgent~\cite{du2024trajagent} unifies various trajectory modeling task through a collaborative agentic framework. Particularly impressive is Coscientist~\cite{boiko2023autonomous}, which autonomously designs, plans, and executes real-world chemical experiments, successfully completing complex palladium-catalyzed cross-couplings. Recently there are also many works on general agent framework, such as AutoGen~\cite{wu2024autogen} and ChatDev~\cite{qian2023chatdev}. 

This trend is particularly evident in traffic simulation domains, where LLM-powered agents can interpret high-level commands and execute corresponding simulator actions. For instance, ChatSUMO~\cite{li2024chatsumo} employs LLMs as a natural language interface to the SUMO simulator, allowing users to issue intuitive commands (e.g., "I want to see traffic in Albany...") that are automatically translated into executable simulator configurations. Similarly, Open-TI~\cite{da2024open} proposes an agentic workflow where users provide natural language instructions through an LLM interface, which then autonomously starts and manages SUMO simulations in the background. These approaches exemplify the broader movement toward LLM-based automation systems that transform abstract user requirements into precise technical operations.

There are also works propose LLM-assisted frameworks to solve traffic signal control (TSC) problems, integrating SUMO for traffic simulation and algorithm validation, such as CityLight~\cite{zeng2024citylight}, LLMAATSC ~\cite{tang2024large}, LLMLight~\cite{lai2025llmlight}, CoLLMLight ~\cite{yuan2025collmlight}, LLM-assisted light~\cite{wang2024llm}, and LLMDUTSC~\cite{tang2024large2}. A more recent and powerful paradigm is the use of Model Context Protocol (MCP)~\cite{anthropic2024mcp}. Frameworks like SUMO-MCP~\cite{ye2025sumo} begin to leverage this concept by wrapping SUMO's functionalities into MCP-compatible tools. 
However, these works are constrained by several key limitations. Firstly, their capabilities are often bounded by pre-defined, rigid workflows, lacking the flexibility to dynamically plan and adapt a comprehensive experimental procedure from scratch. Secondly, their generality is limited: they are often tightly coupled with a narrow set of tasks (e.g., TSC), making it difficult to extend them to a broader range of scenarios. Most critically, they lack a robust mechanism for outcome optimization or fails to enhance global performance, such as minimizing overall travel time, congestion or carbon emission.

\section{Conclusion}
In this paper, we introduced \systemname~, a novel LLM-based multi-agent framework designed to overcome the generality and optimization bottlenecks in traditional traffic simulation workflows. By leveraging a hierarchical architecture of collaborative agents and the carefully designed model context protocol for traffic simulation, \systemname~ translates natural language instructions into executable simulation plans and enables real-time, intelligent optimization of traffic conditions.

Our work makes several key contributions. First, we demonstrate that \systemname~ possesses enhanced generality, capable of handling both specified and ambiguous instructions across diverse online and offline simulation scenarios. Second, we showcase its autonomous optimization capability, where low-level agents, embodying traffic elements, make coordinated decisions that outperform traditional and contemporary LLM-based TSC algorithms. Our analysis on agent reasoning components further shows that good performance stems from a combination of a goal-oriented reward signal and a memory-driven strategy.

Despite its promising results, the current framework has some limitations. The performance and token consumption are inherently tied to the capabilities and scale of the underlying general-purpose LLMs. For future work, we plan to investigate task-specific fine-tuning of more compact language models to reduce computational and token overhead while maintaining performance. 
Furthermore, we will explore to extend the framework to support multi-modal instructions and integrating it with a broader range of urban simulation platforms.

\newpage
\bibliographystyle{ACM-Reference-Format}
\bibliography{reference}

\appendix
\section{Appendix} \label{sec:appendix}
\subsection{Metrics}
\begin{table*}[!t]
\centering
\caption{Evaluation Metrics Used in Experiments}
\label{tab:metrics}
\vspace{-10pt}
\begin{tabularx}{\textwidth}{lllX}
\toprule
\textbf{Category} & \textbf{Metric} & \textbf{Abbrev.} & \textbf{Definition} \\
\midrule

\multirow{6}{*}{\textbf{Efficiency Metrics}} 
& Traffic Volume & TV & Instantaneous count of all vehicles actively traveling in the network at a given simulation step. Reflects real-time traffic load and network occupancy. \\
& Cumulative Throughput & TP & Total number of vehicles that successfully complete their journeys during the entire simulation. Reflects the network’s overall service capacity and operational efficiency. \\
& Estimated Time of Arrival & ETA & Total estimated time required for all departed vehicles to complete their trips under current traffic conditions. Lower values indicate higher system-wide traffic efficiency and reduced delay. \\
& Avg. Travel Time (Finished Vehicles) & ATT-finished & Average time taken by vehicles that successfully complete their trips. Lower values indicate higher individual travel efficiency. \\
& Average Queue Length & AQL & Average number of vehicles waiting in queues at intersections. Lower values indicate less congestion and smoother traffic flow. \\
& ETA Reduction & $\Delta$ETA & Difference between actual travel time and ETA under free-flow conditions. A larger negative value indicates more efficient routing and control. \\

\midrule

\multirow{2}{*}{\textbf{Environmental Impact}} 
& Total Carbon Emission & Carbon & Total CO\textsubscript{2} emissions (in grams) produced by all vehicles during the simulation, calculated based on vehicle attributes (e.g., engine type) and dynamic driving behaviors (e.g., acceleration, speed). Lower values indicate a more environmentally friendly strategy system-wide. \\
& Average Carbon Emission per Vehicle & ACE & Average CO\textsubscript{2} emissions per vehicle (in grams). Lower values indicate a more environmentally friendly strategy on a per-vehicle basis. \\

\bottomrule
\end{tabularx}
\end{table*}

\subsection{Optimization results across different scenarios}
\begin{table*}[h]
  \centering
  \small %
  \caption{Performance comparison of optimization methods across different scenarios.}
  \label{tab:scenario-results}
  \begin{tabular}{lllccccccc}
    \toprule
    Scenario & Sub-scenario & Optimization & \makecell{ATT-f($\downarrow$)} & \makecell{TV($\uparrow$)} & \makecell{$\Delta$ETA($\uparrow$)} & \makecell{ACE($\downarrow$)} & \makecell{ETA($\downarrow$)} & \makecell{TCE($\downarrow$)} & \makecell{TP($\uparrow$)} \\
    \midrule
    \multirow{4}{*}{TSC} 
    & \multirow{2}{*}{evening\_peak} & With & 91.62 & 58 & -146.86 & 0.24 & 62446902.90 & 243.49 & 1006 \\
    & & Without & 383.02 & 428 & -175.99 & 1.10 & 41402628.51 & 733.87 & 665 \\
    \cmidrule(lr){2-10}
    & \multirow{2}{*}{morning\_peak} & With & 133.71 & 68 & -147.26 & 0.22 & 32088206.27 & 283.07 & 1231 \\
    & & Without & 385.32 & 533 & -176.61 & 1.15 & 21344843.20 & 936.34 & 813 \\
    \midrule
    \multirow{4}{*}{Auto-drive}
    & \multirow{2}{*}{female\_dominant} & With & 358.92 & 871 & -173.34 & 1.18 & 59829552.92 & 2106.02 & 1791 \\
    & & Without & 351.19 & 914 & -1175.10 & 1.07 & 58675697.80 & 1885.28 & 1756 \\
    \cmidrule(lr){2-10}
    & \multirow{2}{*}{\makecell{middle\_aged\\middle\_income}} & With & 382.19 & 932 & -161.34 & 1.67 & 51704583.14 & 2889.28 & 1735 \\
    & & Without & 375.54 & 979 & -162.89 & 1.51 & 50560809.29 & 2561.97 & 1696 \\
    \bottomrule
  \end{tabular}
\end{table*}

\begin{table*}[t]
  \centering
  \small
  \caption{Performance comparison of medical service optimization strategies}
  \label{tab:medical-results}
  \vspace{-10pt}
  \resizebox{0.5\textwidth}{!}{%
  \begin{tabular}{lllccc}
    \toprule
    Scenario & Strategy & Optimization & SR (\%) & ATT (s) & Score \\
    \midrule
    \multirow{2}{*}{Medical Service} 
    & \multirow{2}{*}{Mass-benefit} & With & 99.0 & 1735.29 & 0.1725 \\
    \cmidrule(lr){3-6}
    & & Without & 92.5 & 3526.22 & -0.4104 \\
    \midrule
    \multirow{2}{*}{Pediatric Service} 
    & \multirow{2}{*}{--} & With & 100.0 & 2311.29 & 0.0067 \\
    \cmidrule(lr){3-6}
    & & Without & 97.0 & 4251.27 & -0.5963 \\
    \bottomrule
  \end{tabular}}
\end{table*}

\end{document}